%% file: main.tex
\documentclass[10pt,twocolumn,letterpaper]{article}

\usepackage{cvpr}              


\usepackage{graphicx}
\usepackage{url}
\usepackage{booktabs}
\usepackage[textsize=tiny]{todonotes}
\usepackage{makecell}
\usepackage{multirow}
\usepackage{multicol}
\usepackage{tikz}
\usepackage{stfloats}

\definecolor{cvprblue}{rgb}{0.21,0.49,0.74}
\usepackage[pagebackref,breaklinks,colorlinks,allcolors=cvprblue]{hyperref}

\title{Perceptual Inductive Bias Is What You Need Before Contrastive Learning}

\author{
Tianqin Li\thanks{Published in CVPR 2025. Tianqin Li and Junru Zhao contributed equally to this work. Due to a formatting error during the CVPR submission, the equal contribution note was omitted in the official proceedings. This arXiv version corrects that oversight. The author order follows alphabetical order by last name.} \quad 
Junru Zhao\footnotemark[1] \quad 
Dunhan Jiang \quad Shenghao Wu \quad Alan Ramirez \quad Tai Sing Lee \\[1ex]
Carnegie Mellon University \\
{\tt\small \{tianqinl, junruz, dunhanj, shenghaw, alanrami, taislee\}@andrew.cmu.edu}
}

\begin{document}
\maketitle
\input{sec/0_abstract_ready}    
\input{sec/1_intro_ready}
\input{sec/2_related}
\input{sec/3_approach_ready}

\input{sec/4_results_ready}

\input{sec/5_discussion_ready}
\section{Acknowledgement}
This work was supported by NSF CISE RI 1816568 and NIH R01 EY030226-01A1. This work was also supported by the CMU HURAY program.

{
    \small
    \bibliographystyle{ieeenat_fullname}
    \bibliography{main}
}

\end{document}

%% file: sec/0_abstract_ready.tex
\begin{abstract}

David Marr’s seminal theory of human perception stipulates that visual processing is a multi-stage process, prioritizing the derivation of boundary and surface properties before forming semantic object representations. In contrast, contrastive representation learning frameworks typically bypass this explicit multi-stage approach, defining their objective as the direct learning of a semantic representation space for objects. While effective in general contexts, this approach sacrifices the inductive biases of vision, leading to slower convergence speed and learning shortcut resulting in texture bias. In this work, we demonstrate that leveraging Marr’s multi-stage theory—by first constructing boundary and surface-level representations using perceptual constructs from early visual processing stages and subsequently training for object semantics—leads to 2x faster convergence on ResNet18, improved final representations on semantic segmentation, depth estimation, and object recognition, and enhanced robustness and out-of-distribution capability. Together, we propose a pretraining stage before the general contrastive representation pretraining to further enhance the final representation quality and reduce the overall convergence time via inductive bias from human vision systems.


\end{abstract}

%% file: sec/1_intro_ready.tex
\section{Introduction}
\begin{figure}[t]
\begin{center}
\includegraphics[width=0.85\linewidth]{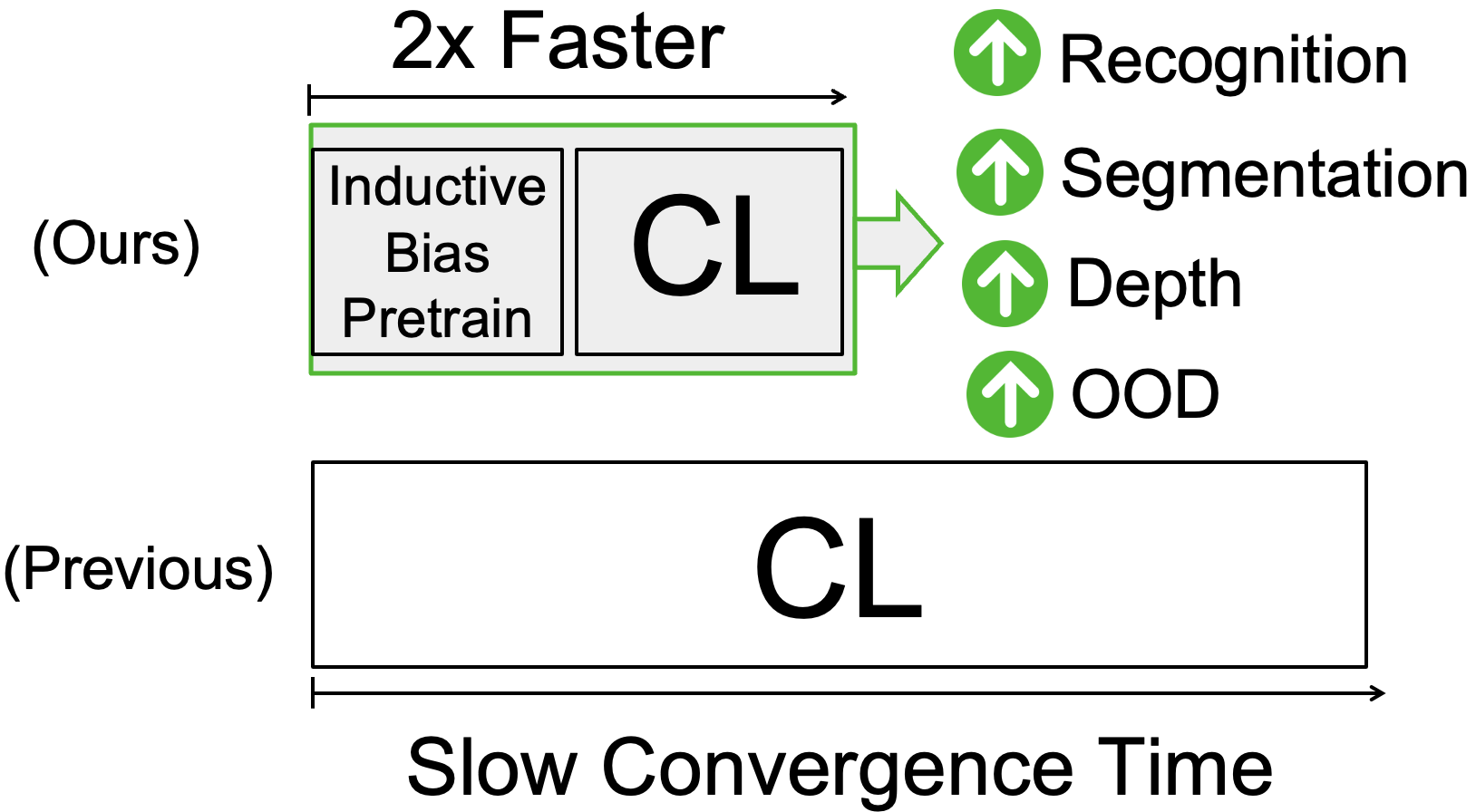}
\caption{We propose to leverage perceptual constructs from early vision processing stages to improve the contrastive learning pipeline. By introducing a pre-pretraining stage before general contrastive learning (CL), we inject inductive bias into the contrastive representation learning. We demonstrate that such inductive bias is beneficial to the representation quality and provides significantly convergence speed-up to the contrastive learning. }
\end{center}
\end{figure}

Human visual information processing is organized in stages, progressing from contrast and edge detection to perceptual organization and figure-ground segmentation, then to computing surfaces and shapes, and ultimately to object recognition. End-to-end supervised learning in neural networks for object classification in images often relies on the simplest features or shortcuts, bypassing essential intermediate representations, resulting in texture bias. Self-supervised contrastive learning, on the other hand, leverages the inductive bias of semantic segregation by grouping semantically similar scenes and separating dissimilar scenes in latent space. This also leads to representations that access rich semantic information without building intermediate representations sensitive to surfaces and boundaries first.

In this work, we evaluate whether perceptual constructs from early visual processing, such as figure-ground segmentation and intrinsic image decomposition, can serve as additional inductive biases for contrastive learning. Evidence of figure-ground organization has been found in the primary visual cortex and secondary visual cortex in primate visual systems \cite{Lamme1995-sz,Zipser7376,LEE19982429, zhou2000coding}. This early perceptual construct distinguishes figures against the ground, assigns border ownership to surfaces, and enables the construction of a neural representation of local shapes and global shapes along the ventral visual hierarchy. The perception of shapes could be innate or at least develop very fast \cite{Wagemans2008-eh,Elder2009-kw}. Furthermore, categorization and identification of unseen objects are considered to rely on the similarity of the overall shape, conceptualized through shape prototypes \cite{shapesimilar}. The recognition of shapes was found to be parallel to the development of vocabularies in language in infancy. Infants can learn new nouns faster when nonce objects are organized in shape, implicating a linked development between labeling objects linguistically and shape prototypes in the visual system \cite{nounlearning}. We conjecture that shape prototypes learned as cluster means of shape silhouettes obtained from figure-ground segregation can allow contrastive learning to learn better intermediate representations for a variety of visual tasks, such as object recognition, segmentation as well as depth estimation in the context of this paper. 

Another set of early perceptual constructs is called intrinsic images \cite{barrow1978recovering}. Intrinsic image decomposition \cite{adelson1996perception, grosse2009ground} is a generalization of Retinex developed by Edwin Land \cite{land1977retinex} to explain color constancy by discounting illumination effects to infer the reflectance or color of the objects. Intrinsic image decomposition factorizes an image as a product of a reflectance image and a shading image. Reflectance images reflect surface material properties and govern our perception of color and lightness. Shading images reflect the interaction of illumination and the 3D surface shape and are thus rich in implicit 2.5D information and affect our depth perception. What kinds of visual tasks would benefit from intermediate representations learned under the guidance of the inductive bias of these intrinsic images?

This work shows that global shapes of figures, particularly through a set of remembered prototypes, provide an inductive bias that enhances contrastive learning, resulting in learned representations that improve object recognition, segmentation, and depth estimation, with an increase in the shape bias of the model. We find that the inductive bias from the shape prototypes increases the speed of representational learning at the beginning, as in the early development of human infants, but ultimately has to be disabled to allow the learning of more refined representations to accommodate a greater variety of semantics. Thus, we propose a hybrid coarse-to-fine inductive bias that allows contrastive learning to learn intermediate representations first with the help of the prototypes and then refine the representation based on semantics. This results in a 2x faster representation learning. On the other hand, intrinsic reflectance images enhance the learned representations for segmentation and object recognition but do not benefit depth estimation. Conversely, intrinsic shading images improve depth estimation capabilities but with little impact on segmentation and object recognition. Finally, we demonstrate that integrating these three perceptual constructs provides additional performance benefits. Thus, we demonstrate that inductive biases derived from early perceptual constructs can differentially or collectively promote the contrastive learning of intermediate representations with greater sensitivity to surface depths and object shapes and better performance in downstream visual tasks.

%% file: sec/2_related.tex
\section{Related Works}
Contrastive learning is one of the most commonly used frameworks in discriminative self-supervised learning. It inherently aligns with the Infomax principle \cite{infomax} by aiming to maximize the mutual information between semantically similar scenes, which are usually different views of the same image. This aim is realized by maximizing InfoNCE loss function \cite{oord2019representation,wu2018unsupervised}. InfoNCE is designed to maximize a lower bound on the mutual information between representations of positive view pairs while distinguishing them from negative view pairs. The views are usually generated by different data augmentations. Many studies \citep{he2020momentum,chen2020simple} emphasize the importance of data augmentations as contrastive learning requires intensive data augmentation to learn good semantic representations. Some works combine contrastive learning with clustering \cite{caron2021unsupervised,PCL}. In this kind of framework, the concept of prototype is introduced to organize the representation space in order to have more compact semantic clusters.

After self-supervised pretraining, the representations are commonly evaluated by downstream tasks such as classification, segmentation and depth estimation. Many works \cite{wang2021dense, o2020dense, yuan2023densedino} point out that there exists a trade-off between image-level classification (semantics) and pixel-wise classification or segmentation (perceptual grouping) in discriminative models. In other words, representations learned by state-of-the-art discriminative models often fail to perform effectively across both high-level semantic tasks and intermediate-level perceptual tasks.

While humans rely heavily on intermediate perceptual constructs like object shapes in object recognition, texture bias is a well-known misalignment between CNN-based models and humans \cite{geirhos2022imagenettrained,NEURIPS2021_c8877cff, geirhos2020surprising}. Many works endeavor to solve this misalignment by different approaches including 1) using data augmentation methods to vary texture, e.g. texture synthesis, style transfer \cite{geirhos2022imagenettrained, wen2023does, geirhos2021partial}, 2) changing to vision transformer \cite{naseer2021intriguingpropertiesvisiontransformers}, or 3) using more complex multimodal generative frameworks \cite{gavrikov2024visionlanguagemodelstexture, jaini2024intriguingpropertiesgenerativeclassifiers}.Improved shape bias results in improvement of recognition robustness and domain generalization \cite{geirhos2022imagenettrained,li2023emergence}. However, a recent study \cite{jaini2024intriguingpropertiesgenerativeclassifiers} argues that most discriminative models are still strongly texture-biased and less robust to tremendous style changes. We attribute this phenomenon to the focus of discriminative models on learning semantic representations while overlooking intermediate-level representations.


%% file: sec/3_approach_ready.tex
\section{Approach}
In this part, we will introduce how we incorporate global shape of figures in a prototypical manner and the intrinsic images as views into contrastive learning.

\begin{figure}[h]
\begin{center}
\includegraphics[width=0.9\linewidth]{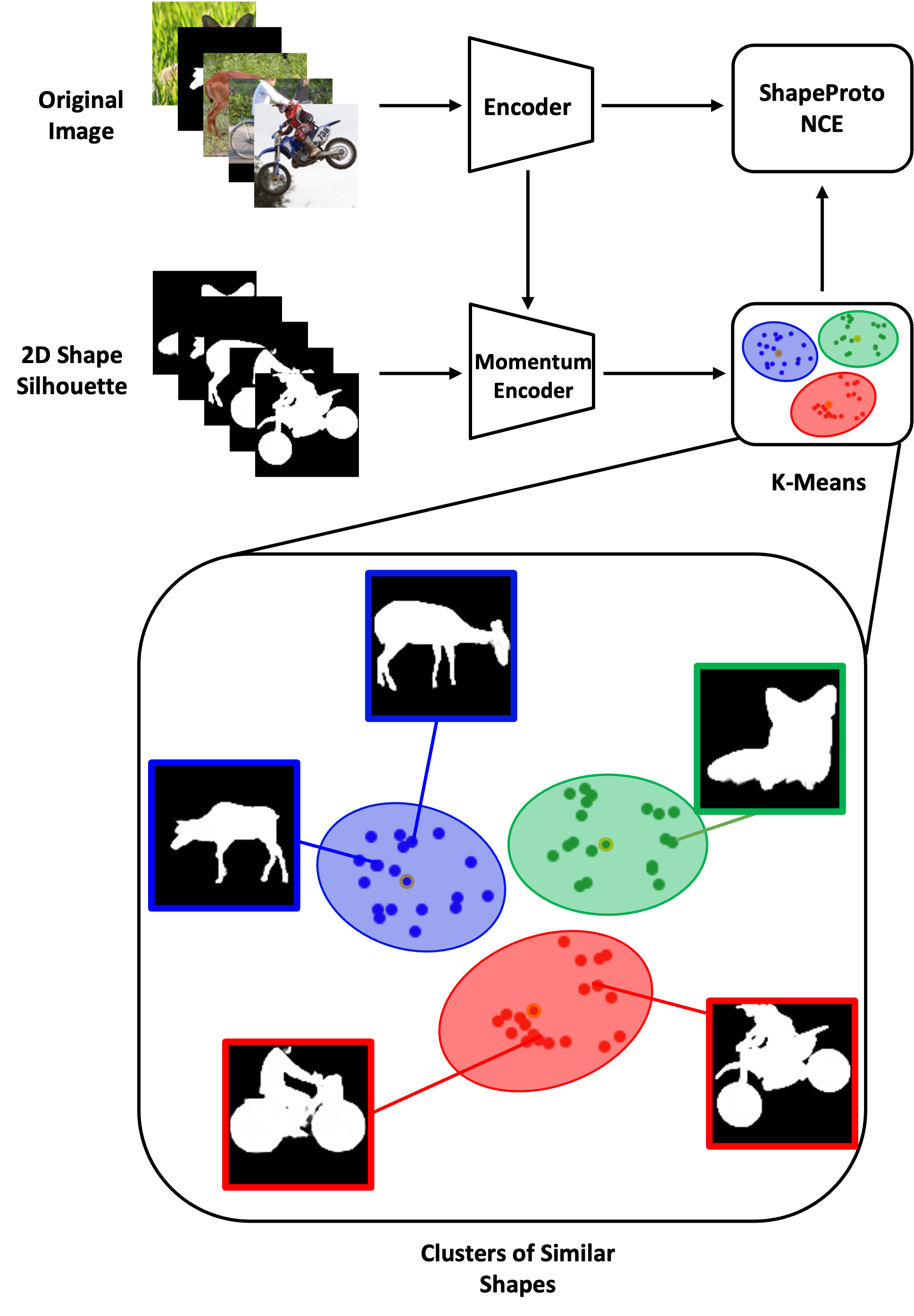}
\caption{\textbf{Illustration of Shape Prototypical Contrastive Learning (S-PCL) training framework} The original images are input
to the encoder and the offline-generated shape silhouettes generated
are input to the momentum encoder. The K-Means is performed
on the embedding of shape silhouettes. Similar shapes are clustered together and the centroid of each cluster is considered to be
the shape prototype. The objective function, ShapeProtoNCE, is
calculated on the representation of the original image and its corresponding shape prototype.}
\label{fig:shape-pcl}
\end{center}
\end{figure}

\paragraph{Shape prototypical contrastive learning (S-PCL)}
\begin{figure}[h]
\begin{center}
\includegraphics[width=0.9\linewidth]{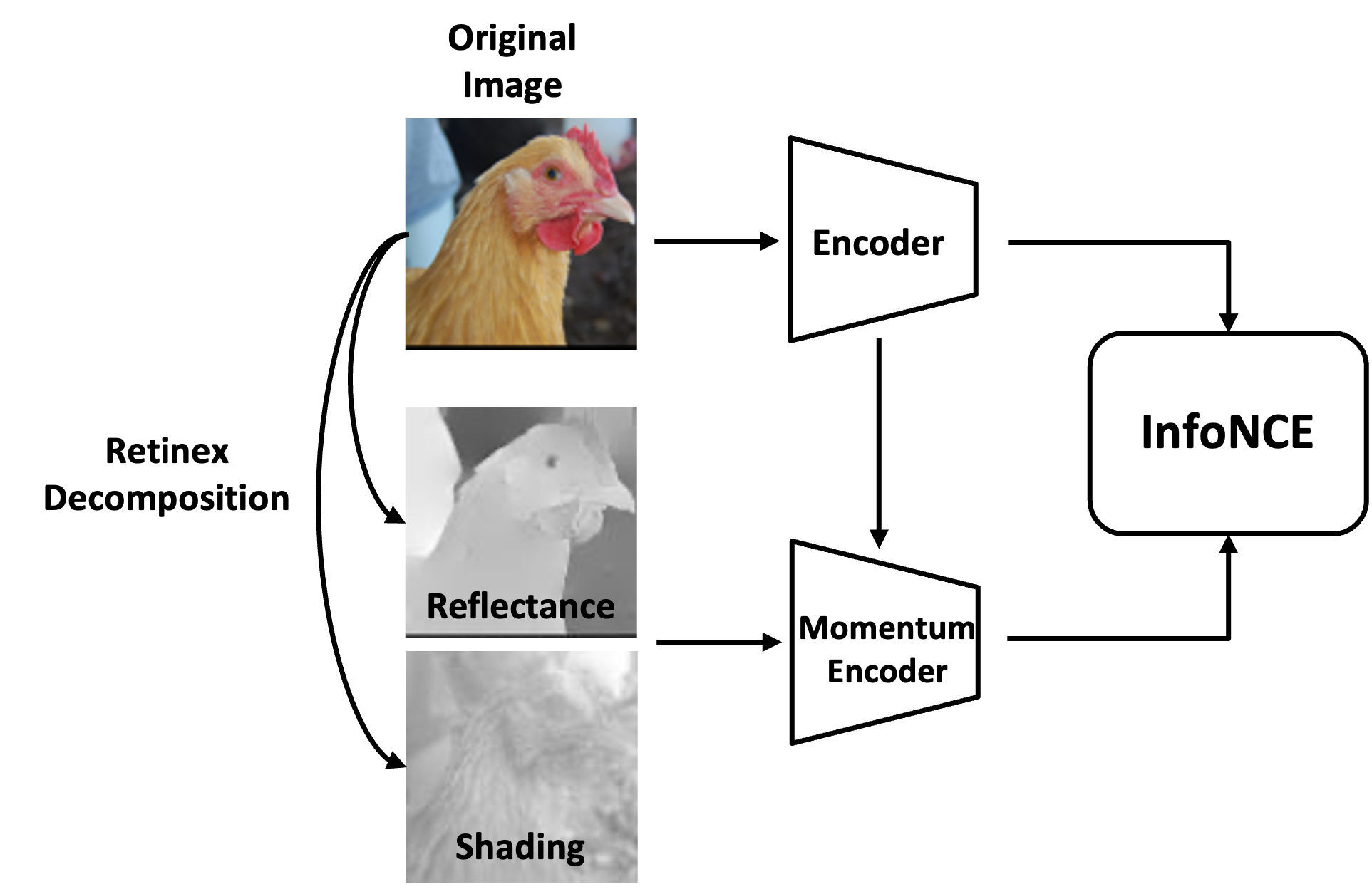}
\caption{\textbf{Illustration of the training framework that incorporates the intrinsic image as a view} The original image is input to the encoder and one intrinsic image decomposed by Retinex algorithm is input to the momentum encoder. The InfoNCE between the representation of the original image and that of the intrinsic image is computed.}
\label{fig:intrin}
\end{center}
\end{figure}
Inspired by the fact that humans rely on the similarity of the overall shape envelope for object recognition, we design a contrastive learning framework that is capable of learning the concept of a group of similar shapes, i.e. shape prototypes. To represent the global shape of an object, we use the figure-ground shape silhouette as presented in Figure.\ref{fig:shape-pcl}. 
Our framework is built upon prototypical contrastive learning (PCL)\cite{PCL}, which involves an online k-means clustering on the representation of the entire training data and considers the centroids of clusters as prototypes. In shape prototypical contrastive learning, the clustering is performed on the representation of shape silhouettes, as illustrated in Figure.\ref{fig:shape-pcl}. The representation learning is guided by the clusters of shapes. With a set of images $X=\{x_1,x_2,...,x_n\}$ and matched shape silhouettes $M=\{m_1,m_2,...,m_n\}$, they are encoded by the same encoder to get their representations $V=\{v_1,v_2,...,v_n\}$ and $U=\{u_1,u_2,...,u_n\}$, respectively. K-Means clustering is employed on the set $U$, producing $K$ clusters and a set of centroids, $S=\{s_1,s_2,...,s_K\}$. The centroid of the cluster is representative of elements in the cluster, and thus is considered as the shape prototype. The mutual information between the representation of the original image $v_i$ and the representation of the shape prototype $s_i$ is maximized by optimizing the InfoNCE objective function:
\begin{equation}
    \mathcal{L}_{\text{ShapeProtoNCE}}=\sum^{n}_{i=1}-\text{log}\frac{\text{exp}(v_i\cdot s_p)/\phi_p) }{\sum^{r}_{j=0}\text{exp}(v_i\cdot s_j/\phi_j)}
\end{equation}
where $s_p$ is the prototype of the cluster $p$ to which $u_i$ belongs, $s_j, j\neq p$ are shape prototypes representing other shape clusters, and $\phi_j$ is the concentration estimation of cluster $j$. Similar to PCL, we perform clustering for $N$ times with different $K_i, i=1,2,...,N$ and average the losses. The final loss function used is the combination of InfoNCE between the normally augmented view and the shape loss:
\begin{equation}
    \mathcal{L}=\mathcal{L}_{\text{InfoNCE}}+\frac{1}{N} \sum_{i=1}^{N}\mathcal{L}_{\text{ShapeProtoNCE},K_i}
\end{equation}

\paragraph{Intrinsic image as a view in contrastive learning}
Unlike shape prototypes, we treat the intrinsic image, either reflectance or shading, as an augmented view. We maximize the mutual information between the representations of normally augmented views and those of reflectance or shading images by optimizing the InfoNCE objective function, as illustrated in Figure \ref{fig:intrin}. The final loss function is the sum of the standard InfoNCE loss between representations of two augmented views and the InfoNCE loss between representations of one augmented view and one intrinsic image. The approaches that incorporate reflectance or shading intrinsic images are termed Reflectance Contrastive Learning (\textbf{ReflCL}) and Shading Contrastive Learning (\textbf{ShadCL}), respectively.

\paragraph{Combination of shape and intrinsic images} We also integrate the shape and the intrinsic image together according to equation (3). We call the combined approach mid-level vision contrastive learning (\textbf{MidVCL}).
\begin{equation}
    \mathcal{L}=\mathcal{L}_{\text{InfoNCE}}+\frac{1}{N} \sum_{i=1}^{N}\mathcal{L}_{\text{ShapeProtoNCE},K_i} + 
    \alpha\mathcal{L}_{\text{Shad}} +
    \beta\mathcal{L}_{\text{Refl}}
\end{equation}
where $\alpha$ and $\beta$ is the weight of the shading and reflectance losses, respectively.
\vspace{-1mm}

\paragraph{Implementation details}
The experiments are conducted on the dataset ImageNet-100\footnote{The dataset used here is reorganized by adjusting the train-validation split to a 7:3 ratio to intensify the challenge}, a subset of ImageNet-1K \cite{deng2009imagenet}, and STL-10 \citep{pmlr-v15-coates11a}. The silhouette mask of the object is generated offline using TRACER \citep{lee2022tracer}. We decompose a full-detail image to a reflectance intrinsic image and a shading image by Retinex algorithm \cite{adelson1996perception, grosse2009ground}. The ResNet18 \citep{He_2016_CVPR} is adopted as the encoder and the output dimension of the last fully-connected layer is 256-D. We follow the data augmentation methods in previous works \cite{chen2020simple, he2020momentum, PCL}. These data augmentations are applied to only the original image and the preprocessing for the shape silhouette and the intrinsic image only contains resizing and normalization. More details can be found in the supplementary material.

%% file: sec/4_results_ready.tex
\section{Results}
\begin{figure*}[h]
  \centering
   \includegraphics[width=\textwidth]{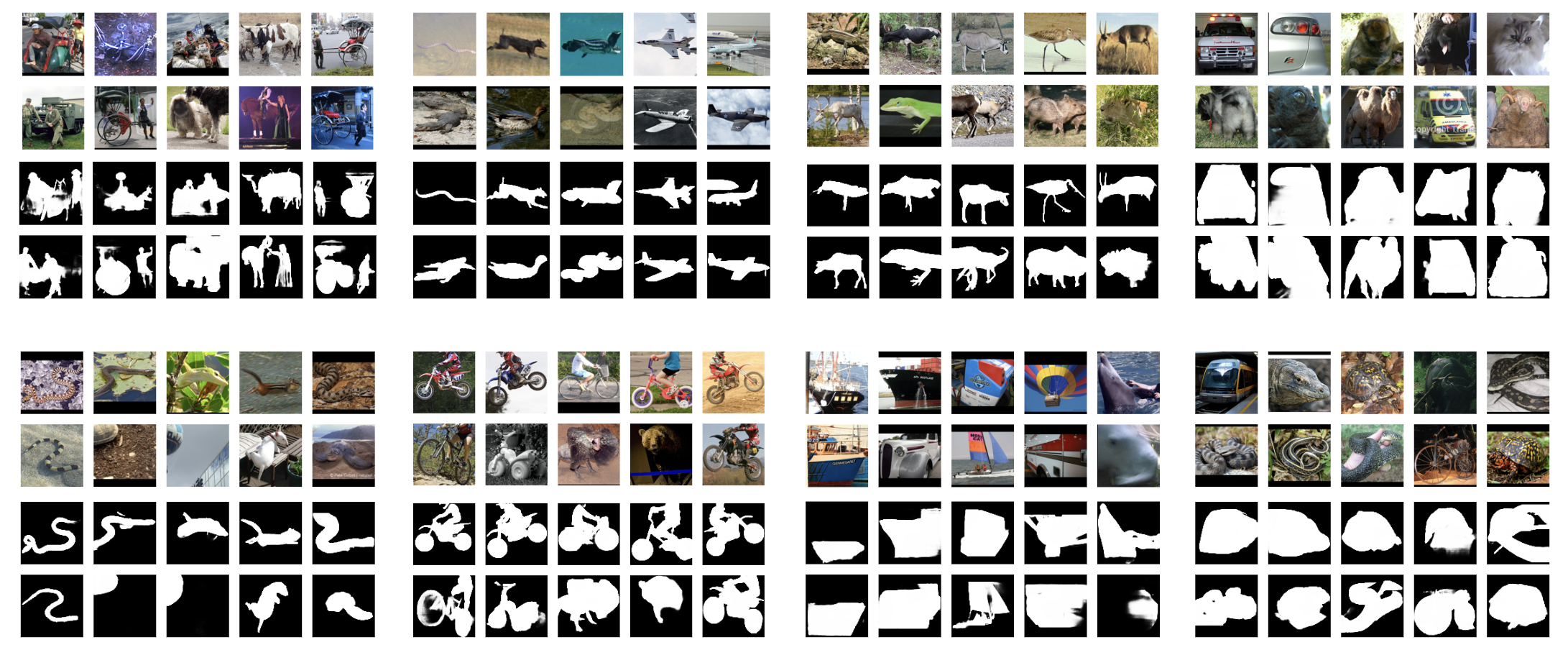}
   \caption{\textbf{Visualization of elements in clusters} produced by S-PCL after trained on STL10 for 400 epochs. Each block represents a cluster and 10 randomly selected images (Top) and shape silhouettes (bottom) are presented. Images in each cluster have similar shape irrespective to their ground truth category, indicating the S-PCL's capability of grouping objects with similar shape.}
   \label{fig:clus_vis}
\end{figure*}

\begin{figure*}[t]
  \centering
   \includegraphics[width=\textwidth]{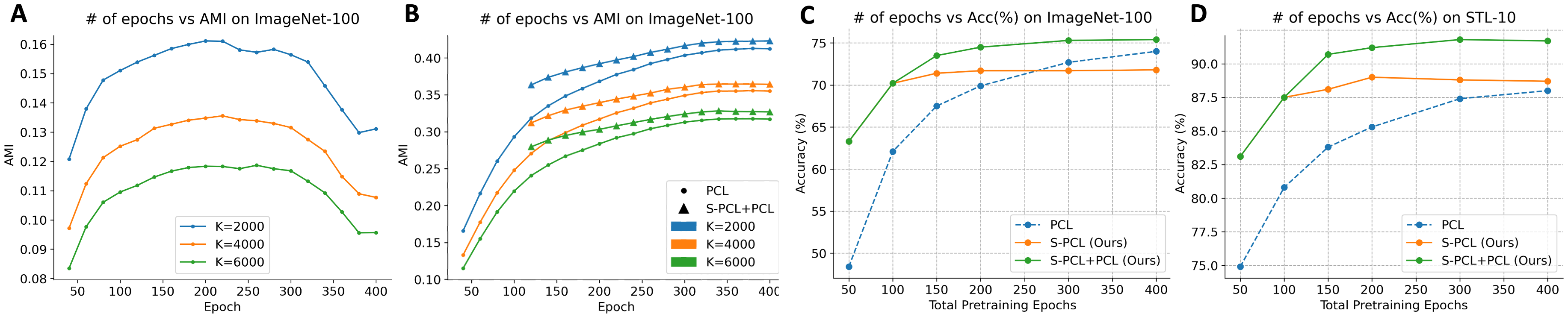}
   \caption{\textbf{A}. AMI between cluster assignment of S-PCL trained on ImageNet-100 and ground truth label. The AMI increases at the early stage and decreases later on. \textbf{B}. AMI between cluster assignment of PCL and S-PCl+PCL trained on ImageNet-100 and ground truth label. The AMI of PCL keeps increasing. S-PCL warm-up moves the curve upward. \textbf{C and D}. Linear classification accuracy on two datasets for PCL, S-PCL and S-PCL+PCL. S-PCL suffers rapid saturation but is able to serve as a warm-up strategy to accelerate training.}
   \label{fig:ami_accele}
\end{figure*}

\subsection{Representation Evaluation}
First and foremost, we evaluate representation learned under the guidance of intermediate perceptual constructs through multiple downstream visual tasks, including linear classification, segmentation, and depth estimation. 
\vspace{-3mm}
\paragraph{Linear classification and transfer learning to ImageNet-1k}
We trained a linear classifier on frozen representations using the entire labeled training set of ImageNet-100 and STL-10, and here we report the Top-1 and Top-5 accuracies of ImageNet-1k on the validation sets. The results on STL-10 are included in the supplementary material. As illustrated in Table.\ref{tab:lin_trans}, S-PCL significantly outperforms the baselines during the early training stages (e.g., at 100 epochs) on both datasets. However, when training is extended to 400 epochs, the performance gains of S-PCL diminish, and it is even surpassed by the baseline approaches significantly on ImageNet-100, indicating a potential faster saturation of S-PCL compared to others. In contrast, ReflCL maintains consistent performance improvements between 100 and 400 epochs, suggesting that reflectance information effectively facilitates object recognition. Conversely, ShadCL, with inductive bias from shading, provides only a tiny gain during the early stages of training and may hinder object recognition when trained for a large number of epochs. MidVCL is similar to S-PCL on the classification task, being the best at epoch 100 while plateauing when the number of epoch increases to 400. We also test generalizability by transferring to a different dataset. This is performed by the linear classification on ImageNet-1k which has 10x more classes than ImageNet-100. The transfer learning results reported in Table.\ref{tab:lin_trans} display the similar trend as the linear classification.

\begin{table}[h]
\setlength{\tabcolsep}{0.4em}
\caption{Linear classification and ImageNet-1k transfer learning accuracies (\%) of different approaches trained for 100 and 400 epochs on the ImageNet-100. For hybrid methods (denoted as "X+Y"), training consists of 100 epochs using method X followed by 300 epochs using method Y. The best-performing results are highlighted in bold.}
\label{tab:lin_trans}
\vskip 0.15in
\begin{center}
\begin{scriptsize}
\begin{tabular}{lccccccr}
\toprule
 & \multirow{2}{*}{\makecell{Epochs}} &\multicolumn{2}{c}{\makecell{ImageNet-100}} & \multicolumn{2}{c}{\makecell{ImageNet-1k}} \\
 &  & \makecell{Top-1\\Acc (\%)$\uparrow$} & \makecell{Top-5\\Acc (\%)$\uparrow$} & \makecell{Top-1\\Acc (\%)$\uparrow$} & \makecell{Top-5\\Acc (\%)$\uparrow$}\\
 \midrule
\makecell{\text{SimCLR}}        & 100 & 65.1 & 87.3  & 33.1 & 58.2 \\
\makecell{\text{MoCoV2}}        & 100 & 61.7 & 86.0  & 31.1 & 55.1 \\
\makecell{\text{PCL}}           & 100 & 62.1 & 86.3  & 30.5 & 54.7 \\
\makecell{\text{SwAV}}          & 100 & 64.8 & 88.3  & 33.4 & 57.8 \\
\makecell{\text{BYOL}}          & 100 & 65.8 & 88.5  & 33.0 & 55.4 \\
\makecell{\text{DINO}}          & 100 & 65.3 & 88.3  & 33.2 & 55.5 \\
\makecell{\text{S-PCL(Ours)}}   & 100 & 70.2 & 90.3  & \textbf{37.2} & 61.9 \\
\makecell{\text{ReflCL(Ours)}}  & 100 & 65.5 & 88.1  & 35.2 & 60.1 \\
\makecell{\text{ShadCL(Ours)}}  & 100 & 62.5 & 86.5  & 32.1 & 55.9 \\
\makecell{\text{MidVCL(Ours)}}  & 100 & \textbf{70.5} & \textbf{90.4}  & 37.0 & \textbf{62.7} \\
\midrule
\makecell{\text{SimCLR}}        & 400 & 77.2 & 93.8  & 40.8 & 66.9 \\
\makecell{\text{MoCoV2}}        & 400 & 77.0 & 93.8  & 41.6 & 68.1 \\
\makecell{\text{PCL}}           & 400 & 74.0 & 91.9  & 37.8 & 63.3 \\
\makecell{\text{SwAV}}          & 400 & 76.8 & 93.6  & 38.3 & 64.2 \\
\makecell{\text{BYOL}}          & 400 & 75.8 & 92.6  & 42.9 & 67.6 \\
\makecell{\text{DINO}}          & 400 & 77.1 & 93.6  & 38.0 & 62.9 \\
\makecell{\text{S-PCL(Ours)}}   & 400 & 71.8 & 90.9  & 37.0 & 63.1 \\
\makecell{\text{ReflCL(Ours)}}  & 400 & 77.2 & 93.4  & 40.0 & 65.4 \\
\makecell{\text{ShadCL(Ours)}}  & 400 & 71.8 & 90.9  & 34.9 & 60.6 \\
\makecell{\text{MidVCL(Ours)}}  & 400 & 72.4 & 90.9  & 37.1 & 61.7 \\
\makecell{\text{S-PCL+MoCoV2(Ours)}}  & 400 & \textbf{78.0} & \textbf{94.2} & \textbf{43.9} & \textbf{69.4} \\
\makecell{\text{MidVCL+MoCoV2(Ours)}}  & 400 & 77.8 & \textbf{94.2} & 43.8 & \textbf{69.4} \\
\bottomrule
\end{tabular}
\end{scriptsize}
\end{center}
\vskip -0.1in
\end{table}

\paragraph{Segmentaion and Depth Estimation}
Besides linear classification, which is a semantic and image-level task, it is also essential to evaluate representations on pixel-level tasks such as segmentation and depth estimation. Therefore, we conduct experiments on various benchmark datasets to assess performance in those tasks. The models evaluated in this part are pretrained for 400 epochs.

For segmentation tasks, we fine-tune the pretrained models with a segmentation head on the ADE20K and Cityscapes datasets. The mean Intersection over Union (mIoU) scores are reported in Table \ref{segmentation}. Our MidVCL approach outperforms all other methods on both datasets, and the S-PCL is on the second place. This superior performance is attributable to the incorporation of shape silhouette masks in both MidVCL and S-PCL, which provide figure-ground information ciritical for segmentation tasks. Additionally, we observe that ReflCL ranks the third, assumably because the reflectance intrinsic image segregates instances based on albedo, thereby facilitating more accurate boundary detection. Also, it turns out that ReflCL performs poorly on depth estimation, suggesting that solely reflectance is insufficient to distinguish foreground and background although the boundaries of instances are clearer. In contrast, ShadCL ends up with only nuanced improvements in segmentation performance.

For depth estimation, we fine-tune the pretrained models with an additional head using the NYU Depth V2 dataset. The relative errors (REL) and root mean errors (RME) are presented in Table \ref{depth}. MidVCL and S-PCL continue to achieve the highest performance among all evaluated approaches. Notably, ShadCL ranks third, indicating that shading information is beneficial for depth estimation tasks. However, the advantages of incorporating reflectance information in this context appear to be limited.\\

\vspace{-3mm}
\noindent \textbf{Overall} Our evaluations on these downstream tasks demonstrate that shape prototypes, reflectance intrinsic image, and shading intrinsic image facilitate representation learning in different ways. The shape prototype helps all three downstream tasks, especially superior in pixel-level tasks (segmentation and depth estimation). However, it is only beneficial in the early stage for the image-level classification task and saturates rapidly. Additionally, the reflectance intrinsic image helps classification and segmentation but not depth estimation. On the other hand, the shading intrinsic image enhances depth estimation thanks to its implicit information of 3D shape and illuminance, but not classification and segmentation, where spatial information is not that important. Among three perceptual constructs, we find the shape prototype yields the most interesting results, thus we conduct further studies to understand the underlying mechanism of the shape prototype.
\begin{table}[h]
\setlength{\tabcolsep}{0.6em}
\caption{Segmentation results. The mIoU(\%) values on ADE20K and Cityscapes are reported. The best performances are in bold.}
\label{segmentation}
\vskip 0.15in
\begin{center}
\begin{footnotesize}
\begin{tabular}{lccccr}
\toprule
 & Epochs & \makecell{ADE20K\\mIoU(\%)$\uparrow$} & \makecell{Cityscapes\\mIoU(\%)$\uparrow$} \\
 \midrule
 \makecell{\text{SimCLR}}  & 400 & 30.4 & 63.4 \\
 \makecell{\text{MoCoV2}}  & 400 & 30.4 & 63.1 \\
\makecell{\text{PCL}}  & 400 & 30.5 & 64.1 \\
 \makecell{\text{SwAV}}  & 400 & 30.2 & 61.9 \\
 \makecell{\text{BYOL}}  & 400 & 30.9 & 61.6 \\
\makecell{\text{DINO}}  & 400 & 30.2 & 61.0 \\
\midrule
\makecell{\text{S-PCL}(Ours)}  & 400 & 31.8 & \textbf{68.3} \\
\makecell{\text{ReflCL}(Ours)}  & 400 & 31.8 & 65.5 \\
\makecell{\text{ShadCL}(Ours)}  & 400 & 31.0 & 64.4 \\
\makecell{\text{MidVCL}(Ours)}  & 400 & \textbf{32.1} & \textbf{68.3} \\
\bottomrule
\end{tabular}
\end{footnotesize}
\end{center}
\vskip -0.1in
\end{table}

\begin{table}[h]
\setlength{\tabcolsep}{0.8em}
\caption{Depth estimation results. The relative error and root mean error on NYU Depth V2 dataset are reported. The best performances are in bold.}
\label{depth}
\vskip 0.15in
\begin{center}
\begin{footnotesize}
\begin{tabular}{lccccccr}
\toprule
 & Epochs & REL$\downarrow$ & RME$\downarrow$ \\
 \midrule
 \makecell{\text{SimCLR}}  & 400 & 0.2545 & 0.1420 \\
 \makecell{\text{MoCoV2}}  & 400 & 0.2550 & 0.1434 \\
\makecell{\text{PCL}}  & 400 & 0.2587 & 0.1437 \\
\makecell{\text{SwAV}}  & 400 & 0.2586 & 0.1459 \\
 \makecell{\text{BYOL}}  & 400 & 0.2543 & 0.1458 \\
\makecell{\text{DINO}}  & 400 & 0.2595 & 0.1451 \\
\midrule
\makecell{\text{S-PCL}(Ours)}  & 400 & 0.2428 & 0.1381 \\
\makecell{\text{ReflCL}(Ours)}  & 400 & 0.2533 & 0.1420 \\
\makecell{\text{ShadCL}(Ours)}  & 400 & 0.2450 & 0.1399 \\
\makecell{\text{MidVCL}(Ours)}  & 400 & \textbf{0.2406} & \textbf{0.1354} \\
\bottomrule
\end{tabular}
\end{footnotesize}
\end{center}
\vskip -0.1in
\end{table}

\begin{table}[th]
\setlength{\tabcolsep}{0.4em}
\caption{Results of linear classification and segmentation of hybrid models. For hybrid methods (denoted as "X+Y"), training consists of 100 epochs using method X followed by 300 epochs using method Y. We can observe that constructing representation with perceptual inductive bias first will boost the performance of contrastive representation learning on both segmentation and classification tasks. The best two performances are in bold.}
\label{hybrid_results}
\vskip 0.15in
\begin{center}
\begin{scriptsize}
\begin{tabular}{lcccccr}
\toprule
 & Epochs & \makecell{ImageNet-100\\Top-1 Acc(\%)$\uparrow$} & \makecell{ADE20K\\mIoU(\%)$\uparrow$} & \makecell{Depth\\RME(\%)$\downarrow$} \\
 \midrule
\makecell{\text{PCL}}  & 400 & 74.0 & 30.5 & 0.1437 \\
\makecell{\text{MoCoV2}} & 400 & 77.0 & 30.4 & 0.1434\\
\midrule
\makecell{\text{S-PCL}(Ours)}  & 400 & 71.8 & 31.8 & 0.1381 \\
\makecell{\text{S-PCL+PCL}(Ours)}  & 400 & 76.0 & 31.9 & \textbf{0.1373}\\
\makecell{\text{S-PCL+MoCoV2}(Ours)}  & 400 & \textbf{78.0} & 31.9 & 0.1398 \\
\midrule
\makecell{\text{MidVCL(Ours)}}  & 400 & 74.0 & \textbf{32.1} & \textbf{0.1354} \\
\makecell{\text{MidVCL+MoCoV2}(Ours)}  & 400 & \textbf{77.9} & \textbf{32.0} & 0.1409 \\
\bottomrule
\end{tabular}
\end{scriptsize}
\end{center}
\vskip -0.1in
\end{table}

\subsection{Cluster Analysis}
\paragraph{Clusters of similar shapes}To gain a deeper understanding of shape prototypes influence the representation learning, we first analyze the clusters produced by S-PCL. As illustrated in Figure.\ref{fig:clus_vis}, we visualize STL-10 images randomly selected from high-density clusters after 400 training epochs. It is evident that the elements within each cluster exhibit similar global shapes, demonstrating the S-PCL's ability to capture shape features and utilize shape as a structural regularity to organize the representation space.
\vspace{-3mm}
\paragraph{Cluster assignment diverges from semantic category in late stages} To study the early superior performance and fast saturation of S-PCL observed in the linear classification evaluation, we analyze the Adjusted Mutual Information (AMI) between K-Means cluster assignments of S-PCL and the categorical ground truth. A higher AMI indicates greater alignment between cluster assignments and semantic categories. As illustrated in Figure.\ref{fig:ami_accele}A, the AMI increases during the initial training stages but progressively decreases as the number of epochs increases. This trend suggests that clustering based solely on shape leads to a divergence from the semantic categories. In other words, it suggests that relying exclusively on shape is insufficient for effective object classification, which greatly requires semantic information. However, the shape shows enormous advantages at the early stage. These findings motivate the development of a hybrid framework that employs S-PCL and MidVCL as a warm-up strategy in the early training stages, which is detailed in \S4.4.

\subsection{Perceptual inductive bias is beneficial for both image-level task and pixel-level task}

\paragraph{Accelerated representation learning for classification}In the results of linear classification (\S4.1) and cluster analysis (\S4.2), we observe tremendous benefits of incorporating shape prototypes. We attribute this phenomenon to an analogy with the accelerated learning of new nouns grouped by shapes during human infancy. Consequently, we investigate a hybrid framework that combines perceptual inductive bias and semantic representation training. Specifically, we propose training with S-PCL or MidVCL for the first 100 epochs and then switching to traditional contrastive learning frameworks (denoted as S-PCL+X and MidVCL+X respectively), allowing the model to discover additional structural regularities and develop a better representation space. Figure.\ref{fig:ami_accele}B shows the AMI results of PCL and S-PCL+PCL. It can be observed that the S-PCL warm-up moves the AMI curve upward. Figure.\ref{fig:ami_accele}C and D show the linear classification accuracies for PCL and S-PCL+PCL on ImageNet-100 and STL10 datasets versus the number of pretraining epochs. For a fair comparison, all frameworks are trained using identical parameter settings and learning rate schedules. The results demonstrate that the S-PCL+PCL method achieves accuracy levels comparable to PCL while requiring less than \textbf{half} the number of training epochs on both ImageNet-100 and STL-10 datasets. The curve of hybrid frameworks of MidVCL is included in supplementary material and it demonstrates the same trend. Overall, our experimental results demonstrate that the sequential learning of intermediate shape representations followed by high-level semantic representations is beneficial for semantic tasks, particularly in accelerating the learning process.
\vspace{-3mm}
\paragraph{Sequential training resolves the trade-off between image-level and pixel-level tasks} Beyond the classification task, hybrid frameworks demonstrate their superiority also in segmentation task. As reported in Table.\ref{hybrid_results}, hybrid frameworks achieve strong performance in both classification and segmentation tasks, unlike methods relying only on either semantic or perceptual representations. Although hybrid frameworks are not as good as S-PCL and MidVCL in depth estimation, they still significantly outperform baseline methods. These results indicate that the addtional learning with perceptual inductive bias before semantic contrastive learning provides a solution to the trade-off mentioned in related works \S2.

\subsection{Concentration on Object Contour and Shape Bias}
\begin{figure}[h]
  \centering
   \includegraphics[width=\linewidth]{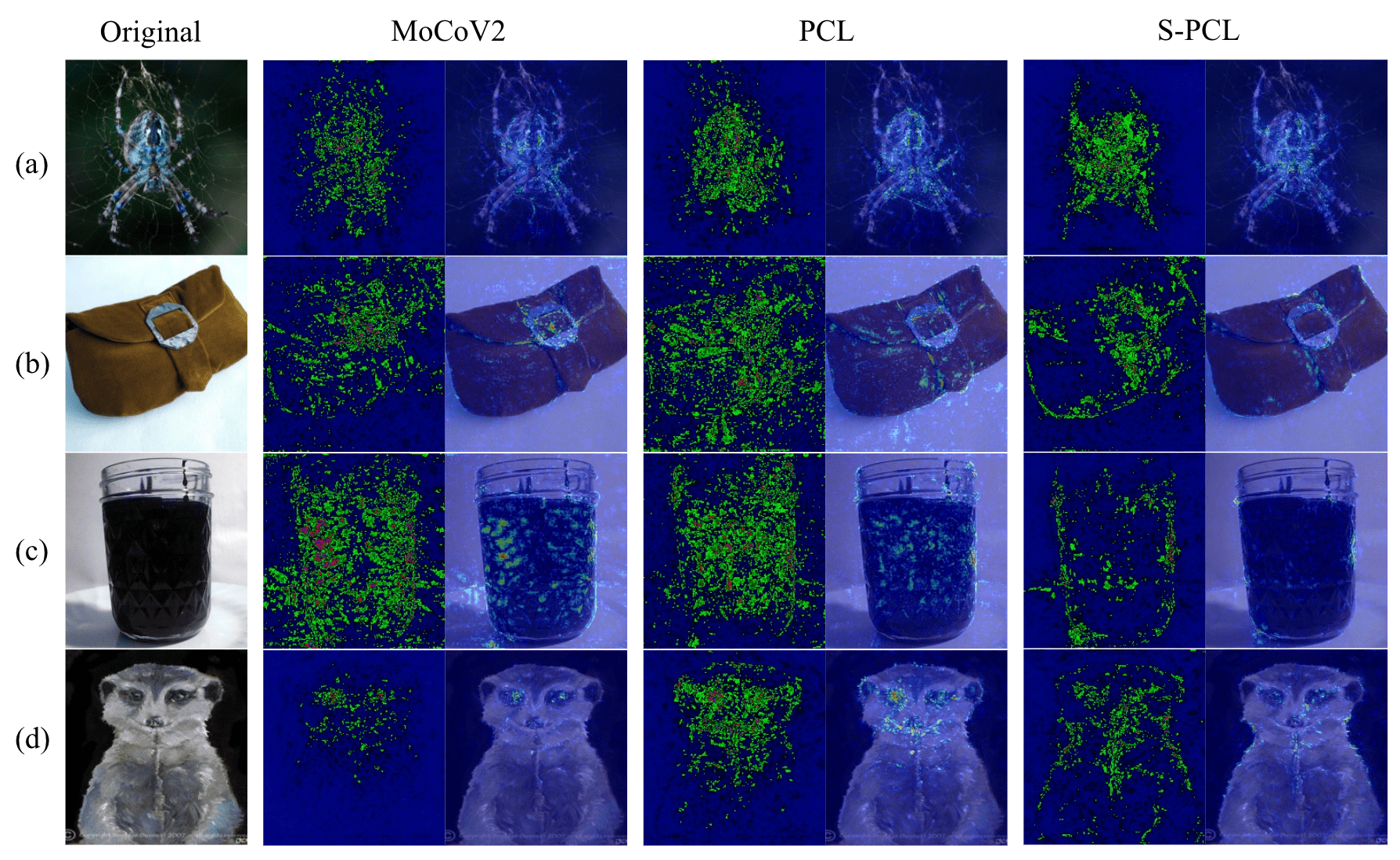}
   \caption{Smoothgrad sensitivity on different images.}
   \label{fig:sen_map_0}
\end{figure}

\paragraph{Shape silhouette forces model to focus on object contours}Next, we study how shape representations vary the model's prediction decision-making process. To this end, we employed the SmoothGrad method to generate sensitivity maps, where highlighted regions contribute significantly to the model's inferences. We show some example images in the validation set of ImageNet-100 in Figure.\ref{fig:sen_map_0}. Notably, comparing the sensitivity maps of S-PCL and others, we find that the semantic representations lead to a stronger concentration on local features of the object, which is a shortcut for classification. While the shape representations guide the model to focus on the contour of the object. 

\begin{figure}[h]
  \centering
   \includegraphics[width=\linewidth]{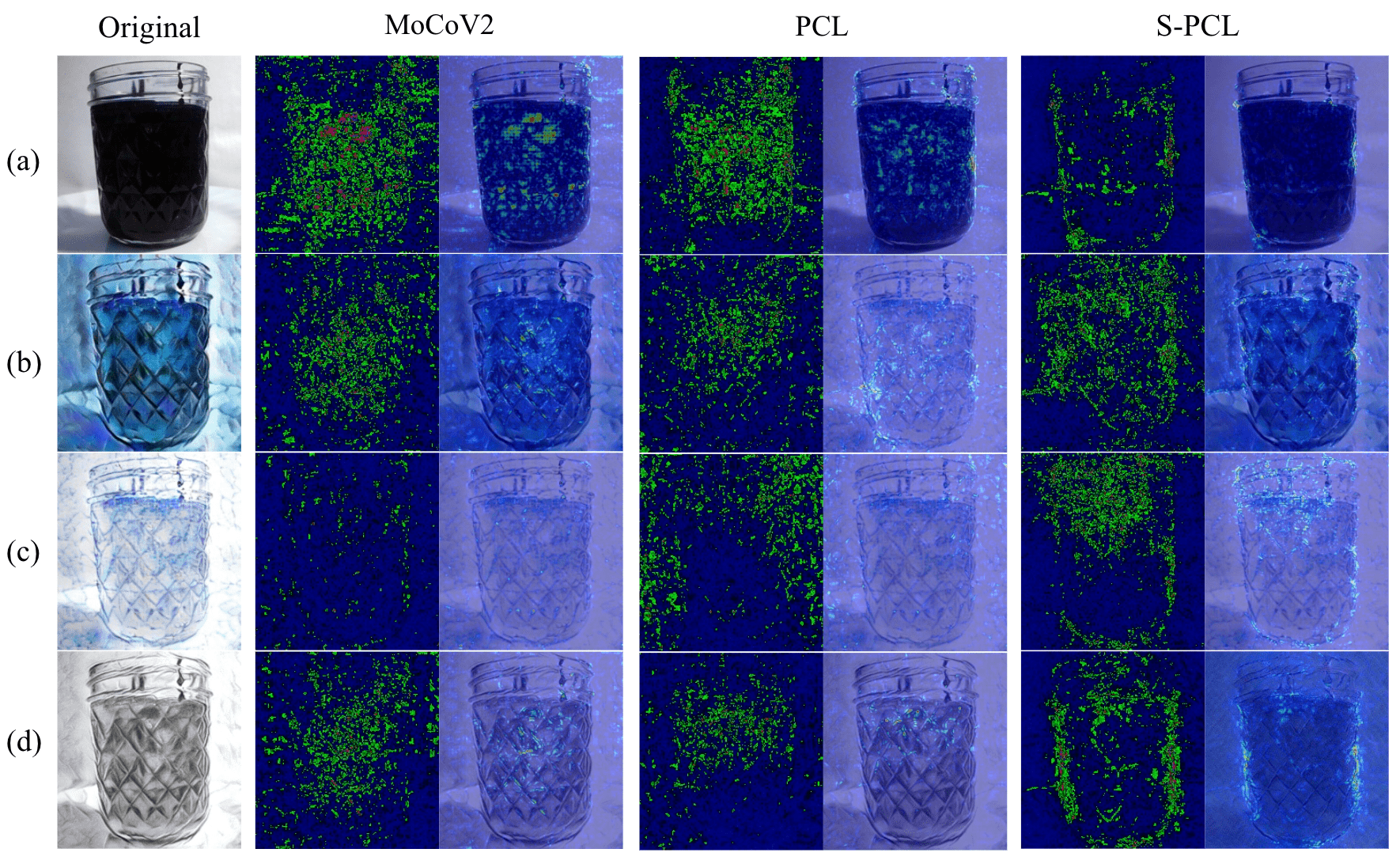}
   \caption{Smoothgrad sensitivity map on the image in different styles.}
   \label{fig:sen_map_1}
\end{figure}

One may argue that although MoCoV2 and PCL focus more on texture, the silhouettes of objects can still be observed to some degree. To address this, we also analyze sensitivity maps of style-transferred images, that is, preserving object shape while altering texture, in addition to the original images, see Figure.\ref{fig:sen_map_1}. After enormous style changes, MoCov2 and PCL are detracted by those irrelevant local features while S-PCL still has a strong concentration on the contours of the objects.

\begin{figure}[h]
  \centering
   \includegraphics[width=\linewidth]{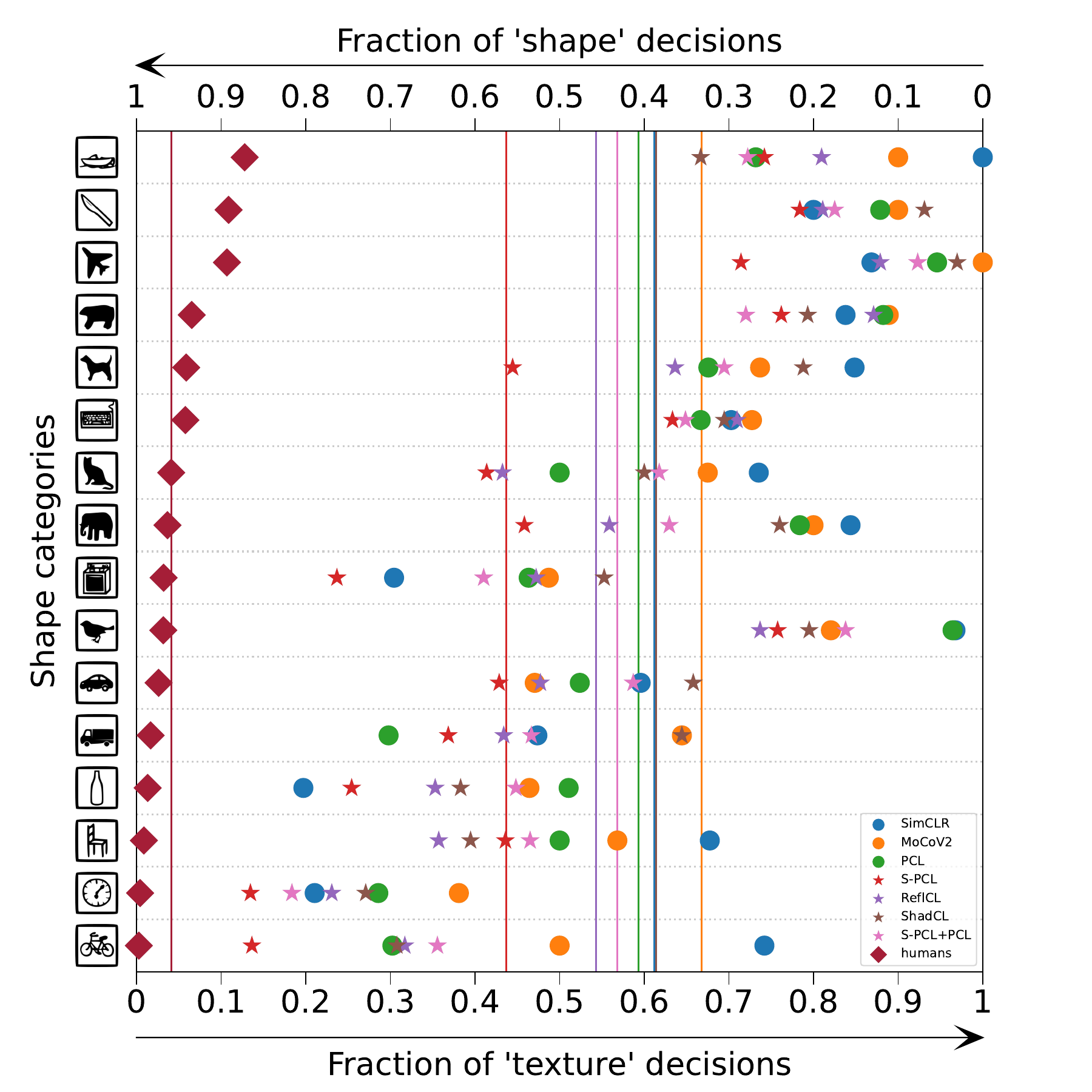}
   \caption{\textbf{The shape bias score.} The value on the top represents the fraction of shape decisions and the value at the bottom represents the fraction of texture decisions. Two values sum to 1 and 0.5 means neutrality, indicating no bias towards either shape or texture. Markers are the fraction of certain shape categories and vertical lines are the mean fraction across shape categories. The star markers represent our proposed approaches.}
   \label{fig:shape_bias}
\end{figure}
\begin{figure}[h]
  \centering
   \includegraphics[width=\linewidth]{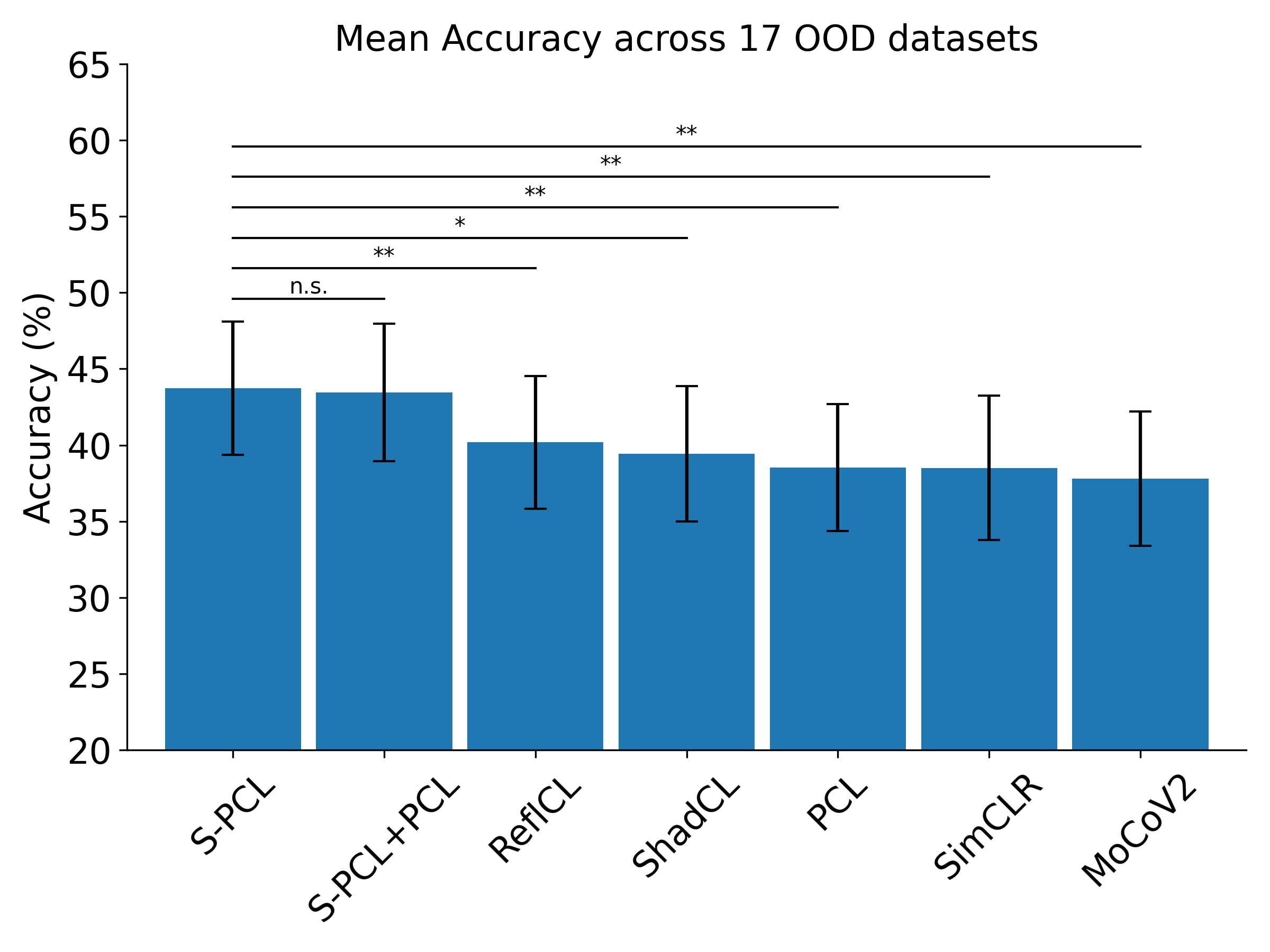}
   \caption{\textbf{Mean accuracy }across 17 OOD dataset in model-vs-human benchmark. Bars are positioned in the descending order. The error bar is the standard error of the mean (SEM). A paired T-test is conducted. ** denotes $p<0.01$, * denotes $p<0.05$, n.s. denotes ``not significant" $p>0.05$.}
   \label{fig:ood}
\end{figure}

\vspace{-3mm}
\paragraph{Shape bias score}
So far, we have found that the shape prototype can increase the shape bias of the model, alleviating a well-known machine's misalignment with humans. To quantitatively evaluate the shape bias of the model, we test our model on the model-vs-human benchmark. The model-vs-human benchmark is a protocol that is able to evaluate the 2D shape bias of a model. With a set of images with conflict cues, e.g. a cat shape with elephant skin texture, we can know which is the predominant visual cue for object recognition between shape and texture. Figure.\ref{fig:shape_bias} demonstrates shape score across different frameworks. Importantly, the proposed SPCL is the only model that uses shape as the predominant cue, i.e. shape bias score greater than 0.5. RelfCL shows improvements compared to the rest of the approaches but it is still biased towards texture (fraction of shape decisions less than 0.5).
\vspace{-3mm}
\paragraph{Out-of-distribution robustness}
Many studies reveal that the model with high shape bias will also have higher robustness to out-of-distribution (OOD) data. To this end, we test models on the 17 OOD datasets included in the model-vs-human benchmarks. Figure.\ref{fig:ood} presents average classification accuracies across all 17 OOD datasets in the model-vs-human benchmark. The full results are included in the supplementary material. S-PCL demonstrates statistically significant improvements in OOD accuracy compared to baseline models. This suggests that leveraging shape information not only aids in better generalization on seen data but also enhances the model's performance on unfamiliar inputs.

%% file: sec/5_discussion_ready.tex
\section{Discussion}

In this work, we aim to integrate inductive biases from early and intermediate perceptual constructs into contrastive learning rather than relying solely on high-level semantic information. To achieve this, we maximize the mutual information between full-detail images and one or more perceptual constructs. Our findings reveal that inductive bias from global shape prototypes not only accelerates early-stage classification but also improves segmentation and depth estimation while enhancing the model’s overall shape bias. Reflectance intrinsic images facilitate representation learning for classification and segmentation, although they do not benefit depth estimation, whereas shading intrinsic images enhance depth estimation without aiding classification or segmentation. Combining all three perceptual constructs enables us to leverage the unique advantages of each. Moreover, employing sequential training—where the model first harnesses perceptual inductive biases before applying semantic contrastive learning—effectively mitigates the trade-off between image-level and pixel-level tasks. Overall, our results indicate that incorporating perceptual constructs generated by early and intermediate visual processing into self-supervised learning can help networks develop distinct intermediate representations that improve downstream visual tasks, offering a promising avenue for future research in contrastive learning.

\section{Limitation}
We developed distinct approaches for specific perceptual constructs, focusing on shape, figure-ground, and intrinsic images. These constructs were chosen due to their close association with fundamental visual tasks like classification, segmentation, and depth estimation. However, additional mid-level perceptual constructs merit further exploration. Moreover, our experiments were conducted on a small model and limited datasets because of computing power constraints, underscoring the need to address scalability in future research.